\pdfoutput=1

\documentclass[11pt]{article}

\usepackage{acl}

\usepackage{times}
\usepackage{latexsym}

\usepackage[T1]{fontenc}

\usepackage[utf8]{inputenc}

\usepackage{microtype}
\usepackage{amssymb}
\usepackage{amsopn}
\usepackage{graphicx}
\usepackage{multirow}
\usepackage{makecell}
\usepackage[english]{babel}
\usepackage{bm}
\usepackage{array}
\usepackage{setspace}
\usepackage{CJKutf8}
\usepackage{algorithmicx}  
\usepackage{algpseudocode}  
\usepackage{amsmath}  
\usepackage{color,xcolor}
\usepackage{algorithm} 

\usepackage{pifont}
\usepackage{bm}
\usepackage{amsfonts}

\usepackage{graphicx}
\usepackage[normalem]{ulem}
\usepackage{multirow}
\usepackage{amsmath}
\usepackage{url}

\usepackage{tikz}
\usepackage{xcolor}
\usepackage{tcolorbox}
\usepackage{color,xcolor}

\usepackage{amssymb}
\usepackage{amsmath}
\usepackage{amsopn}
\usepackage{graphicx}
\usepackage[english]{babel}
\usepackage{bm}
\usepackage{array}
\usepackage{setspace}
\usepackage{todonotes}

\usepackage{CJKutf8}
\usepackage{xspace}
\usepackage{amsfonts}
\usepackage{array,multirow}
\usepackage{pgfplots}
\usepackage{tikz}
\usepackage{verbatim}
\usepackage{subfig}
\usepackage{arydshln}
\usepackage{booktabs}
\definecolor{ugreen}{rgb}{0,0.5,0}
\definecolor{mygreen}{RGB}{58,127,88}
\definecolor{iyellow}{RGB}{255,250,205}
\definecolor{ipurple}{RGB}{230,230,250}
\definecolor{myred}{RGB}{160,52,52} 
\definecolor{myblue}{RGB}{30,144,255}
\definecolor{myorange}{RGB}{255,127,80}
\definecolor{mypurple}{RGB}{255,20,147}

\title{Semformer: Transformer Language Models with Semantic Planning}

\author{
 Yongjing Yin$^{1,2}$, Junran Ding$^{2}$, Kai Song$^{4}$, Yue Zhang$^{2,3}\thanks{\ \ Corresponding author}$ \\
 $^1$ Zhejiang University\\
 $^2$ School of Engineering, Westlake University\\
 $^3$ Institute of Advanced Technology, Westlake Institute for Advanced Study\\
 $^4$ ByteDance\\
 \quad{\{yinyongjing,dingjunran\}@westlake.edu.cn} \\
 \quad{yue.zhang@wias.org.cn} \\
}

\begin{document}
\maketitle
\begin{CJK}{UTF8}{gbsn}

\begin{abstract}

Next-token prediction serves as the dominant component in current neural language models.
During the training phase, the model employs teacher forcing, which predicts tokens based on all preceding ground truth tokens.
However, this approach has been found to create shortcuts, utilizing the revealed prefix to spuriously fit future tokens, potentially compromising the accuracy of the next-token predictor.
In this paper, we introduce Semformer, a novel method of training a Transformer language model that explicitly models the semantic planning of response.
Specifically, we incorporate a sequence of planning tokens into the prefix, guiding the planning token representations to predict the latent semantic representations of the response, which are induced by an autoencoder.
In a minimal planning task (i.e., graph path-finding), our model exhibits near-perfect performance and effectively mitigates shortcut learning, a feat that standard training methods and baseline models have been unable to accomplish.
Furthermore, we pretrain Semformer from scratch with 125M parameters, demonstrating its efficacy through measures of perplexity, in-context learning, and fine-tuning on summarization tasks\footnote{https://github.com/ARIES-LM/Semformer.git}.


\end{abstract}

\section{Introduction}
Neural language models (LMs) \cite{bengio-etal-2008-nnlm}, a fundamental component of natural language processing (NLP), have witnessed significant advancements in recent years. 
By scaling up model sizes and pretraining on extensive text, large language models (LLMs) have successfully learned language and world knowledge, which has resulted in promising performance across various tasks and even demonstrated reasoning capabilities \cite{gpt3,Wei22,arxiv2023:openai_gpt4,arxiv2023:LLaMA,DBLP:conf/nips/SchaefferMK23}. 
The success of these models can be attributed to a straightforward training paradigm: next-token prediction with teacher forcing \cite{DBLP:journals/neco/WilliamsZ89}, in which the models are trained to predict tokens using all preceding ground truth tokens as input.

\begin{figure}[t]
\centering
\includegraphics[width=1.0\linewidth]{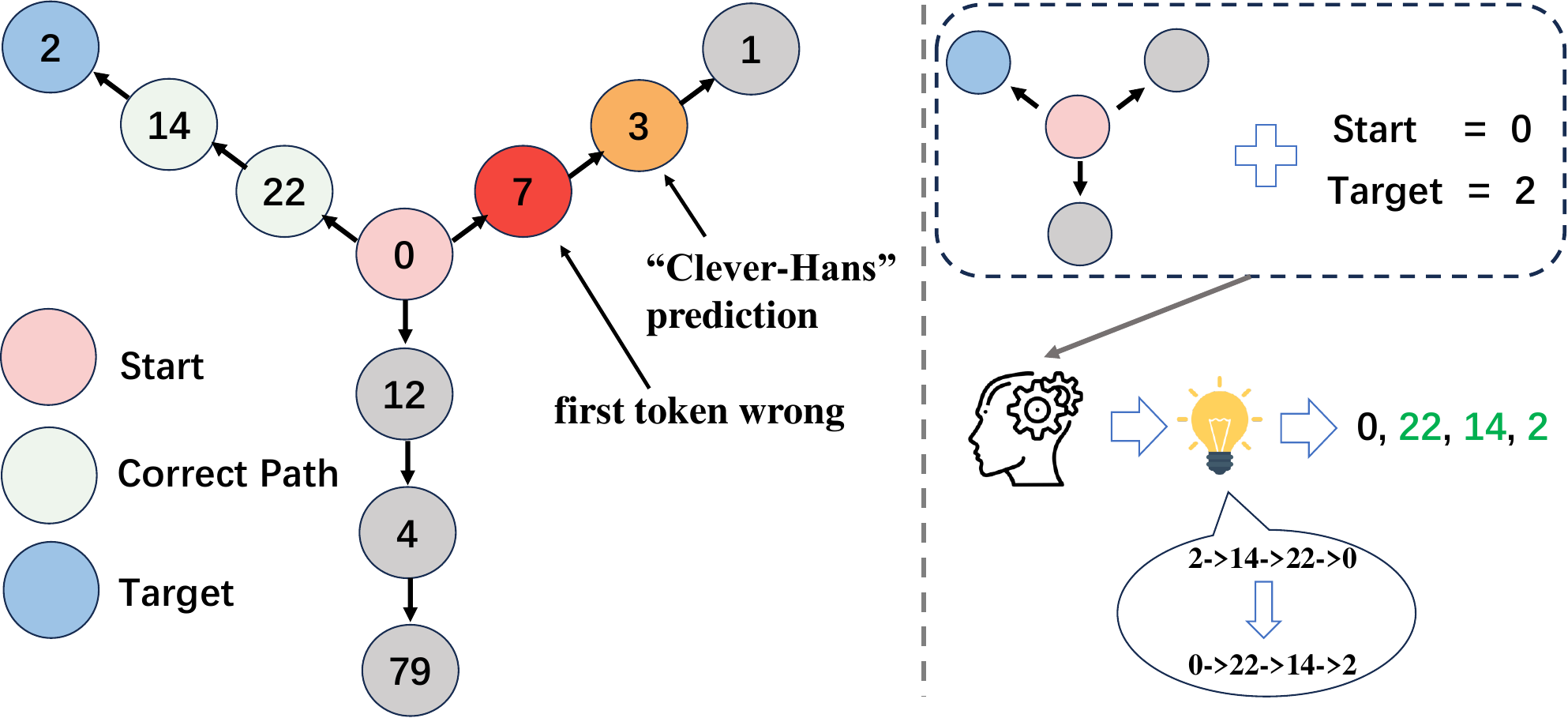}
\caption{
The Clever Hans cheat in a graph path-finding problem which is a minimal lookahead task.
The task is to find the correct path based on the adjacency list, the start node, and the target node.
}\label{fig_intro_cleverhans}
\end{figure}

Recent studies, however, have raised concerns about the efficacy of the aforementioned training scheme in facilitating the learning of an accurate problem solver or planner \cite{pmlr-v235-malach24a,DBLP:conf/iclr/WiesLS23,pmlr-v235-bachmann24a,multitokenpred,DBLP:journals/corr/abs-2404-15758}. 
For instance, the graph path-finding task—which necessitates lookahead and planning— demonstrates that teacher forcing can lead to a Clever Hans Cheat phenomenon characterized by shortcut learning \citep{pmlr-v235-bachmann24a}. 
Consequently, the later nodes such as 3 and 1 in Figure \ref{fig_intro_cleverhans} become easier to predict, while the first node of the answer (i.e., 22) becomes more challenging to learn. 
This could result in a highly inaccurate next-token predictor, which would struggle to generalize to unseen problems, even without considering out-of-distribution and length generalization.


Humans, intuitively, do not rely solely on historical context to solve a problem \cite{DBLP:journals/corr/abs-2305-12272}.
Instead, they formulate an abstract plan based on the problem at hand, which subsequently guides them towards the final answers. 
For the problem in Figure \ref{fig_intro_cleverhans}, the quickest solution is to look ahead at the later nodes to identify a unique path corresponding to the problem, and then reverse this found path to generate the correct answer.
In general, a language model should internalize the process of looking ahead or "thinking about the future".
The semantics of finding the response path is predicted by internal computation, with token output guided by the intended semantics. 


To this end, we incorporate semantic planning into next-token prediction in a decoder-only Transformer \cite{vaswani2017attention,Radford2019LanguageMA}, which we refer to as Semformer. 
Our Semformer is composed of a language model and an autoencoder that is used only during training.
For the language model, we introduce a semantic planning token sequence that follows the prefix of the input. 
This sequence disregards the general next-token prediction loss and is utilized to predict the latent representations of the subsequent tokens. 
The autoencoder learns to generate the sequence of latent representations, compressing the subsequent tokens into a low-dimensional space.


On the graph path-finding problem 
\cite{pmlr-v235-bachmann24a}, our Semformer achieves almost 100\% accuracy scores on the settings of different levels of difficulty, showing superiority to the related baselines. 
Only introducing dummy tokens in the sequence (i.e., Pause Transformer) \cite{DBLP:journals/corr/abs-2310-02226} fails to learn the planning task.
Moreover, our Semformer learns to solve the problem significantly faster than the baselines, merely one epoch based on the GPT2-Large \cite{Radford2019LanguageMA}.
Further, to validate the effectiveness of this architecture on general LM pretraining, we train Transformer models with 125M parameters from scratch on OpenWebText.
Semformer results in improvements on perplexity evaluation, in-context learning, and fine-tuning on abstractive summarization.

\section{Related Work}

\paragraph{Next-token Prediction}
for next-token prediction.
Despite being the standard training objective, next-token prediction has faced several challenges.
On one hand, criticisms target the error accumulation caused by \textit{autoregressive inference}
\cite{kaariainen2006lower,pmlr-v9-ross10a,DBLP:conf/nips/DziriLSLJLWWB0H23,lecun2024}.
On the other hand, there has been debate about whether \textit{teacher forcing} can learn an accurate next-token predictor especially for reasoning and planning tasks.
\citet{DBLP:journals/corr/abs-2303-12712} report failures on GPT4 experimental report and they speculate the failures result from the “linear thinking” in next-token prediction.
\citet{DBLP:journals/corr/abs-2305-12272} informally note that some next-tokens can be hard to learn as they require a global understanding of what will be uttered in the future.
\citet{pmlr-v235-bachmann24a} demonstrate the Clever Hans cheat and the inference error can happen at the beginning. 
While language models are often shown to perform worse on out-of-distribution data
\cite{DBLP:journals/corr/abs-2309-13638},  \citet{pmlr-v235-bachmann24a} demonstrate that they can fail even test in the same distribution.
In addition, \citet{pmlr-v235-malach24a} and \citet{DBLP:conf/iclr/WiesLS23} argue that some complex multi-hop tasks become learnable via next-token prediction only when providing a preceding chain-of-thought supervision for each hop.
\citet{DBLP:journals/corr/abs-2404-15758} also find that the learning of using filler tokens necessitates specific and dense supervision.
The above studies support our motivation to provide general dense supervision for language models.

Beyond the next-token prediction, various training paradigms have been proposed including non-autoregressive models \citep{DBLP:conf/iclr/Gu0XLS18}, diffusion LM \cite{DBLP:conf/nips/LiTGLH22,DBLP:conf/nips/0002G0ZSJ23}, and multiple token prediction \citep{DBLP:conf/emnlp/QiYGLDCZ020,DBLP:journals/corr/abs-2311-13581,multitokenpred}.
Predicting multiple future tokens is originally for accelerating inference, and \citet{multitokenpred} recently show that it can also avoid the localness issue of next-token prediction with teacher forcing.
\citet{DBLP:conf/nips/0002G0ZSJ23} introduce a latent diffusion model to generate paragraph representations induced by a variational autoencoder \cite{DBLP:journals/corr/KingmaW13}, and feed them into the language model to help paragraph generation.
Rather than significantly changing the model architecture, we internalize the planning ability into the language model, achieved through the semantic representation prediction of the subsequent sequence.

\paragraph{Custom Tokens in Language Modeling}
Custom tokens can be used to increase model capacity used as additional memory \cite{DBLP:journals/corr/abs-1907-01470,DBLP:journals/corr/abs-2006-11527,bulatov2022recurrent}.
For example, \citet{bulatov2022recurrent} propose to apply custom tokens recurrently, leading to improvement on long sequence modeling and algorithmic tasks. 
Compressing long prompts into fixed length sequence can alleviate the heavy burden of a large key-value cache during inference \citep{li-etal-2023-compressing,DBLP:journals/corr/abs-2308-08758,DBLP:conf/nips/Mu0G23}.
Custom tokens are also used to optimize pretrained models to accomplish specific downstream tasks, i.e., parameter efficient fine-tuning \citep{Lester-etal-2021-the-power-of,li-liang-2021-prefix}.
For vision Transformer, \citep{DBLP:journals/corr/abs-2309-16588} find that appending trainable tokens to image patches leads to smoother representation learning.

Incorporating trainable tokens has been demonstrated as an effective way to enhance the Transformer's reasoning and planning capabilities.
\citet{herel2024thinking} find that such a method leads to small perplexity gains on reasoning tasks, and \citet{DBLP:journals/corr/abs-2310-02226} investigate its effectiveness on the setting of pretraining on C4 with the evaluation on math and question answering.
\citet{wang2024guiding} propose the addition of new tokens preceding each CoT step.
\citet{zelikman2024quietstar} generate rationales post every token to elucidate future thinking using REINFORCE learning. 
Our work is in line with the above studies in introducing additional tokens.
The differences lie in that our purpose is to alleviate the shortcut learning induced by teacher forcing, and the tokens are used to generate the semantic plan representations of the subsequent tokens.
More importantly, we use a simple and general representation prediction method to guide the function learning of the custom tokens.

\section{Method}

\begin{figure}[!t]
\centering
\includegraphics[width=1.0\linewidth]{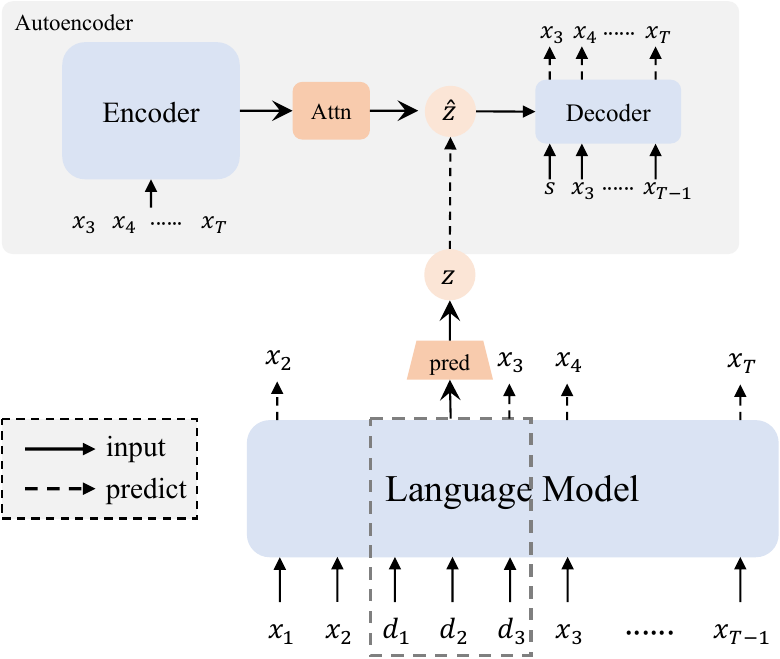}
\caption{
Illustration of our Semformer.
We introduce trainable tokens in language modeling.
The representations of the tokens encoded by the language model are regressed to the latent representations of the response with $L_2$ loss.
We can share the parameters between the language model and the encoder, and utilize a small decoder to enhance training efficiency.
}\label{fig_model_arch}
\end{figure}

\subsection{Next-token Prediction}\label{sec_ntp}

Given an observed text sequence of length $T$, $x=\{x_1,...,x_T\}$, neural language models \cite{bengio-etal-2008-nnlm} (NLM) are trained to predict every token conditioned on the previous tokens defined by the chain rule of probability, i.e., teacher forcing:
\begin{equation}
    \log p_{\theta}({x})=\sum_{t=1}^T\log p_{\theta}(x_t|x_{<t}),
\end{equation}
where $\theta$ is the model parameter.
During inference, the model autoregressively generates the response token-by-token by sampling or searching strategies, given the prefix and all previously generated tokens.
We use decoder-only Transformer as the language model. 
However, our method can also be applied to other architectures such as Mamba \cite{gu2024mamba}.

\subsection{Semformer}

In addition to next-token prediction, we introduce the prediction in the representation space.
The overall framework of our Semformer is illustrated in Figure \ref{fig_model_arch}.
Specifically, we use an autoencoder to learn latent representations of the target sequence, which guides the representation learning of the language model.

During training, we segment each input sequence $x$ into the prefix $x_{1:n}$ and target $x_{n+1:T}$ where $n$ is the segmentation position between the prefix and the response.
For general LM pretraining, the position is selected randomly for each sequence block.
Then, we append $k$ trainable planning tokens $d=\{d_1,d_2,...,d_k\}$ to the prefix. 
The input of the language model can be rewritten to $x' = \{x_{1:n};d;x_{n+1:T}\}$.
We feed $x'$ into the language model and the planning tokens are not used for the loss calculation for predicting the next token.
Formally, the training loss is defined by:
\begin{equation}
    \mathcal{L}_{\text{LM}}=\sum_{t=1 \atop x'_t \notin d}^{T+k}\log p_{\theta}(x'_t|x'_{<t}).
\end{equation}

\paragraph{Latent Semantic Planning}
We provide the plan tokens with generic supervision information, enabling them to serve as the function to compute a future plan before response generation.
The supervision is to predict the latent semantic planning representations of the response, and we introduce an autoencoder with a bottleneck layer to this end.

The encoder of the autoencoder takes the response $x'_{n+1:T}$ as the input and encode them into contextualized representations $H_r$, which are then compressed into a sequence of latent vectors $Z=\{z_1,z_2,...,z_k\}$ using a cross-attention layer:
\begin{align}
    H_{r}&=\text{Encoder}(x'_{n+1:T}), \\
    Z&=\text{CrossAttend}(Q, H_r, H_r),
\end{align}
where $Q$ is the trainable query input of the cross-attention layer.
We use a linear transformation to project $Z$ into a low-dimension representation space.
The number of latent vectors is the same as the number of the planning tokens.
Using cross-attention provides us with more flexible options for the encoder, such as sharing parameters with the language model or using an off-the-shelf pre-trained encoder.

We treat $Z$ as additional memory for the decoder. 
Before being fed into the decoder, each latent vector is projected into the same dimension of hidden states of the decoder with a distinct linear transformation.
Then, the latent vectors are attended by other tokens via self-attention. 
Such an infusion mechanism of latent vectors is convenient to apply pretrained language models without any modification, and has been shown superior to regarding $Z$ as extra input token embeddings \cite{DBLP:conf/emnlp/LiGLPLZG20}.
The objective is the standard reconstruction loss:
\begin{equation}
    \mathcal{L}_{\text{AE}} = \log p_{\theta_{\text{AE}}}(x_{n+1:T}|Z),
\end{equation}
where $\theta_{\text{AE}}$ is the parameter set of the autoencoder.

To alleviate the training burden, we can adopt the following strategies:
sharing the parameters between the encoder and the language model,
using an off-the-shelf encoder,
stopping the gradient flow into the encoder in the autoencoding branch,
and using a compact decoder.

\paragraph{Latent Representation Prediction}
Given the contextualized representations $H$ of the input $x'$ encoded by the language model, we use a predictor head to output the predicted latent representations.
The loss is defined as the $L_2$ distance between the predicted representations and the target latent representations:
\begin{equation}
\mathcal{L}_{\text{RP}} = \sum_{i=1}^k\Vert z_i-f_{\theta_{\text{RP}}}(H_{n+i})\Vert_{2}^2,
\end{equation}
where $f_{\theta_{\text{RP}}}$ is the representation predictor with parameter $\theta_{\text{RP}}$, and we use a linear transformation shared across different positions.

\paragraph{Overall Training Objective}
The whole framework is jointly optimized as follows: 
\begin{equation}
    \mathcal{L}= \mathcal{L}_{\text{LM}} + \mathcal{L}_{\text{AE}} + \alpha \mathcal{L}_{\text{RP}},
\end{equation} 
where $\alpha$ is the coefficient of the latents prediction loss.
By compelling the model to predict the abstract representations of the future response in advance, we can mitigate the Clever Hans cheat issue that arises from exposure to the ground-truth prefix.

\paragraph{Inference} During inference, we simply append the planning tokens to the prefix, and the inference remains standard autoregressive decoding.

\section{Experiments on Graph Path-finding}

\begin{table*}[!t]
\centering
\begin{tabular}{lcccccccc}
\toprule
\textbf{Model} & G(2,20) & G(5,20) & G(5,30) & G(10,20) & G(15,15) & G(20,5) & G(30,5) & G(20,10) \\
\midrule
\multicolumn{9}{c}{GPT2-Large} \\
\midrule
Standard &49.2 &20.1 &19.8 &10.1 & 6.8&4.8 &3.0 &4.9 \\
Teacher-less  & 1.7 &97.8 &0.0 &0.0 &0.0 &99.9 &99,8 & 1.8 \\
Multi-token & 51.0 &19.6 &20.0 &10.1 &6.8 &99.9 &3.3& 4.9 \\
BoW &\textbf{100.0} &\textbf{99.9} &87.9 &85.3 &99.0 &99.9 &99.9 &\textbf{99.9}\\
Pause & 49.9 &20.0 &19.7 &9.7 &6.9 &5.0 &3.2 &4.8 \\
Semformer  &99.9 &\textbf{99.9} &\textbf{99.2} &\textbf{99.6} &\textbf{99.5} & \textbf{100.0} &\textbf{100.0} & \textbf{99.9} \\
\midrule
\multicolumn{9}{c}{GPT2-Small} \\
\midrule
Standard &49.6 &19.7 &19.9 &9.8 &6.7 &4.9 &3.2 &4.8 \\
Teacher-less  &0.0  &0.0 &0.0  & 0.0 &0.0  &5.0 &99.5 & 0.0 \\
Multi-token   & 50.2 & 19.8 & 20.3 & 10.1 & 5.0 & 4.9 & 3.3 & 4.9 \\
BoW           & \textbf{99.9} & 95.1 & 82.7 & 10.3 & 82.3 & 99.9 & \textbf{99.9} & 4.9 \\
Pause  &50.0  & 19.9 & 19.9  & 10.0 & 6.6 &5.0  & 3.3 & 5.0 \\ 
Semformer  & \textbf{99.9} &\textbf{99.5} &\textbf{99.0} &\textbf{98.0} &\textbf{99.1} &\textbf{100.0} &99.6 & \textbf{99.9} \\
\bottomrule
\end{tabular}
\caption{
Accuracies on the graph path-finding test sets.
The setting G($d$,$l$) is characterized by the degree of the node at the center $d$ and the length of each path $l$, respectively.
The number of node values $N$ is the product of $l$ and $d$, omitted for simplicity.
The results for Standard and Teacher-less are obtained by running the code released by \citet{pmlr-v235-bachmann24a}, and the other baselines are re-implemented.
}\label{tab_main_large}
\end{table*}

The graph path-finding task, as introduced by \citet{pmlr-v235-bachmann24a}, involves a unique structure known as a path-star graph $G(d,l,N)$.
Each graph features a central node from which $d$ distinct paths emerge, each comprising $l$ nodes, including the central node. 
The parameter $N$ represents the range of node values, randomly selected from the set $\{0,1,...,N-1\}$, and may exceed the total number of nodes in the graph. 
The input of language models includes all of the edges of the star-graph, the start node, and the end node. 
The objective is to accurately predict the sole correct path between the designated start and end nodes.

In particular, both the training and test graphs are derived from the same distribution, maintaining consistent topology characterized by fixed values of $d$, $l$, and $N$. 
This setup ensures that the observed failures are attributable to in-distribution errors rather than lack of compositional or length generalization capabilities. 
Given that each graph is uniquely labeled and features a randomized adjacency list, the model is required to deduce a general algorithmic solution.
Following \citet{pmlr-v235-bachmann24a}, the dataset comprises 200,000 training samples and 20,000 test samples. 
The number of node values $N$ is set as the product of $l$ and $d$, facilitating a diverse range of graph instantiations.

\subsection{Settings}

\paragraph{Baselines}
We use the pretrained GPT2-Large and GPT2-Small \cite{Radford2019LanguageMA} as the base models of our experiments\footnote{We use the open-source resource at https://github.com/gregorbachmann/Next-Token-Failures.git}. 
We then compare our Semformer with the following baselines:
   (1) \textbf{Standard}, which uses standard teacher forcing training;
   (2)\textbf{Teacher-less}, which predicts multiple future tokens at once (i.e., non-autoregressive generation) \cite{pmlr-v235-bachmann24a};
    (3) \textbf{Multi-token}, which predicts the following multiple tokens using different output heads \cite{multitokenpred};
    (4) \textbf{BoW}, which predicts bag-of-words of the target sequence;
    (5) \textbf{Pause}, \cite{DBLP:journals/corr/abs-2310-02226} which appends planning tokens and learns them only using the language modeling loss.

\paragraph{Hyper-parameters}
We train all models using a batch size of 32 for a maximum of 100 epochs. 
The AdamW optimizer is employed with a learning rate set at 1e-5. 
For Semformer, the number of planning tokens is set to 4 and the coefficient $\alpha$ is set to 1.0 by default.
We use the language model as the encoder and the decoder is set to 6 layers to enhance training efficiency. 
In more challenging configurations, such as G(10,20), while an $\alpha$ of 1.0 remains effective, increasing it to 10.0 significantly accelerates convergence.
For Multi-token, we employ a three-token strategy. For Pause, we insert a number of planning tokens equivalent to those used in Semformer. 
For the BoW approach, we predict the bag-of-words from the average pooled representations of the planning tokens. 
The regularization coefficient for BoW is set to 0.1 through a grid search.

\subsection{Main Results}\label{exp_main}
The evaluation results are presented in Table \ref{tab_main_large}. 
Overall, Semformer achieves near-perfect performance across all the graph configurations.
The standard Transformer encounters significant challenges in learning the planning task accurately, due to the Clever Hans cheat learned by teacher forcing.
In particular, the accuracy for predicting the first node following the start node is approximately $1/d$.
Once the first node after the start node is provided, the model demonstrates a high level of accuracy in generating the entire corresponding path \cite{pmlr-v235-bachmann24a}.

The non-autoregressive Teacher-less models avoid the pitfalls of the cheat to fit the training data.
They demonstrate impressive performance on configurations such as G(5,20), G(20,5), and G(30,5) when using the GPT2-Large. 
However, these models encounter difficulties with the longer responses, which can lead to significant challenges in fitting the training data and results in complete failure (i.e., accuracy 0.0) during test phases. 
The Multi-token approach does not offer particular advantages and only works on G(20,5) with the shortest target path.
The difference between Multi-token and Semformer is that Semformer is trained to predict the complete semantic planning of the target while Multi-token is only trained to predict local future tokens.
In particular, Pause does not learn to solve this problem.
This indicates that simply increasing computing capacity may not be enough to learn lookahead skills effectively, echoing the theoretical research on the competencies of filler tokens \cite{pmlr-v235-malach24a,DBLP:conf/iclr/WiesLS23}.
The BoW method can be regarded as a simplified variant of Semformer.
It disregards the sequence dependency of the target and only considers surface token information.
When integrated with GPT2-Large, BoW achieves commendable results in some settings due to the enforcement of predicting the overall nodes in the target path. 
Nevertheless, it underperforms in scenarios involving longer target sequences, such as G(5,30) and G(10,20).

We also explore the impact of model size by employing GPT2-Small, which is approximately one-sixth the size of GPT2-Large. 
Remarkably, our Semformer still maintains nearly 100\% accuracy scores without modification to the hyperparameters, while the performance of other baseline models declines.
For instance, in configurations such as G(5,30) and G(10,20), the performance of BoW deteriorates to the level of random guessing, exhibiting the underlying limitation in the simple token prediction.

\paragraph{When Semformer Falls Short.}

\begin{table}[!t]
\centering
\begin{tabular}{l|c}
\toprule
Graph & Accuracy \\
\hline
G(21,10)  & 60.2 \\
G(23,10) & 22.1 \\
G(25,10) & 2.8 \\
\midrule
G(40,10) & 99.8 \\
G(10,40) & 10.0 \\
\hline
\end{tabular}
\caption{
\label{tab_fail}
Performance of Semformer under more challenging settings.
}
\end{table}

We further evaluate Semformer under more challenging conditions to identify scenarios where its performance may falter (Table \ref{tab_fail}). 
In fact, \citep{pmlr-v235-bachmann24a} specifically excludes out-of-distribution testing scenarios to better control variables. 
We test on G(21,10), G(23,10), and G(25,10) using the Semformer model trained on G(20,10) to simulate out-of-distribution conditions. 
The results indicate that the model has some extrapolation ability on G(21,10), but fails in more different situations.

Additionally, we conduct experiments on G(40,10) and G(10,40) to assess performance on larger graphs. Since the sequence length in these settings exceeds the maximum length of GPT2 (i.e., 1024), we switch the backbone to pythia-410m\footnote{https://huggingface.co/EleutherAI/pythia-410m}. 
In the case of the same number of nodes, the graph with a longer path is more difficult. 
Our model can achieve 98\% accuracy on G(40,10), but it failed to learn successfully on G(10,40).

\subsection{Analysis}

\subsubsection{Convergence of Different Models}

\begin{figure}[t]
\centering
\includegraphics[width=1.0\linewidth]{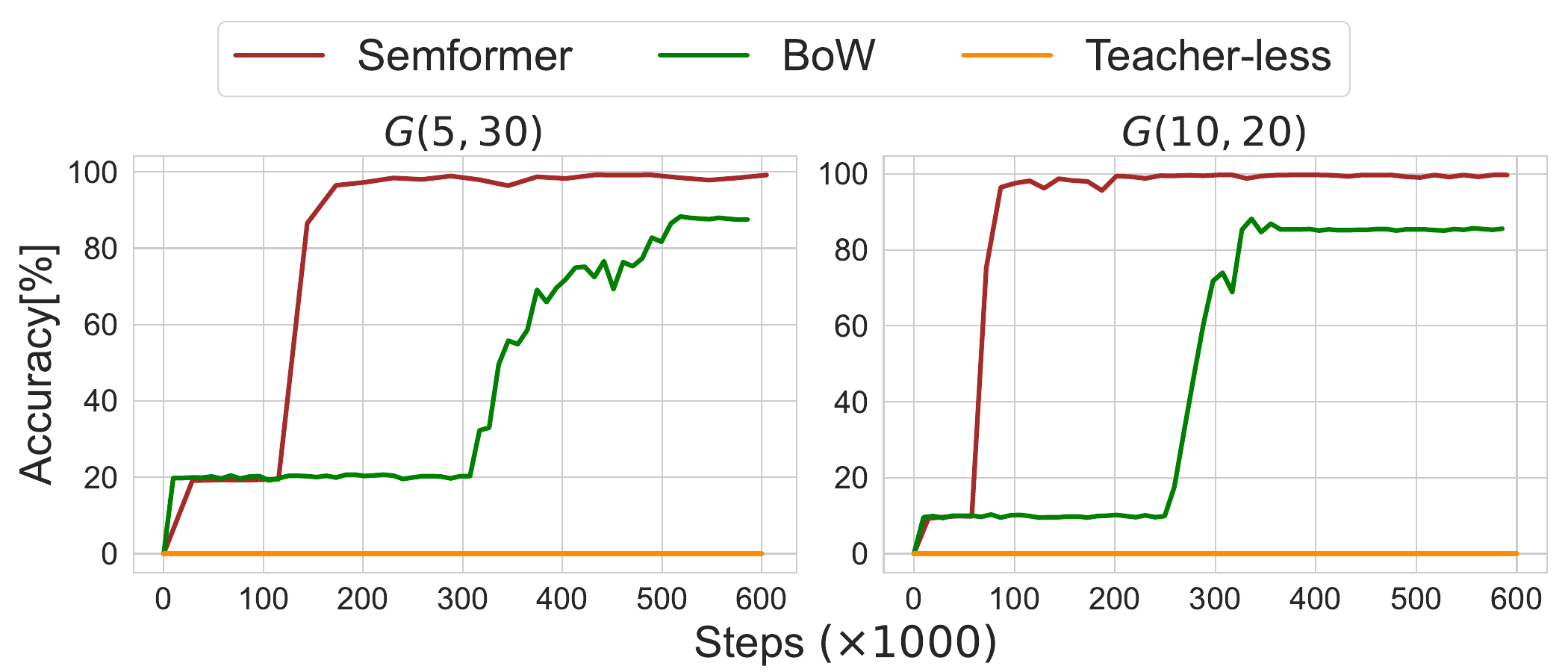}
\caption{
Convergence curves of Teacher-less, BoW, and our Semformer on tasks G(5,30) and G(10,20).
}\label{convergence}
\end{figure}

We choose the graph setting G(5,30) and G(20,10), then display the accuracies with training steps in Figure \ref{convergence}. 
Teacher-less fails on both tasks, yielding an accuracy of 0. 
Semformer achieves peak accuracy in less than 50,000 steps. 
In contrast, BoW requires over 4 times more training steps than Semformer to converge, and fails to attain perfect accuracy on both tasks.
These results demonstrate that our framework provides a highly efficient supervisory signal to learn the lookahead skill.

\subsubsection{Ablations of Autoencoder}

\paragraph{Encoder Design}

\begin{table}[!t]
\centering
\small
\begin{tabular}{l|c|c}
\toprule
Model & 10,20 & 20,10 \\
\hline
NonAE  & 9.6 & 4.7 \\
NonAE(ema) & 9.6 & 5.0 \\
AE & 99.6 & 99.9 \\
\hline
\end{tabular}
\caption{
\label{tab_encoder_design}
Encoder design. 
The use of an autoencoder works better than using the language model itself as the encoder.
}
\end{table}

An alternative method is using the language model itself as the encoder to induce the latent planning representations instead of specially training an autoencoder. 
Concretely, the language model takes the concatenation of the target sequence and the planning tokens as input, and the latents are obtained by stacking a linear transformation on the contextualized representations.
We also attempt to using the exponential moving average trick to generate the encoder, which has shown effectiveness in contrastive learning \cite{he2020momentum}.
The results in Table \ref{tab_encoder_design} demonstrate the advantage of using a separately trained autoencoder, which can learn more meaningful and structured abstract representations than the simple encoding of the input information.

\paragraph{Decoder Layers}
The number of decoder layers in the autoencoder used in the main results is 6, and we further investigate its influence on performance. 
We choose a challenging setting, G(10,20), and use GPT2-Large as the base model. 
For configurations with 1, 3, 6, and 12 layers, the test accuracy scores all exceed 99\% and the model with 6 decoder layers converges slightly faster than the others.
This result is reasonable since a one-layer Transformer decoder can achieve satisfactory performance on language reconstruction \cite{DBLP:conf/emnlp/Montero0S21}, and reconstructing a path sequence is simpler than reconstructing natural language.

\paragraph{Latent Dimension}
\begin{figure}[t]
\centering
\includegraphics[width=0.9\linewidth]{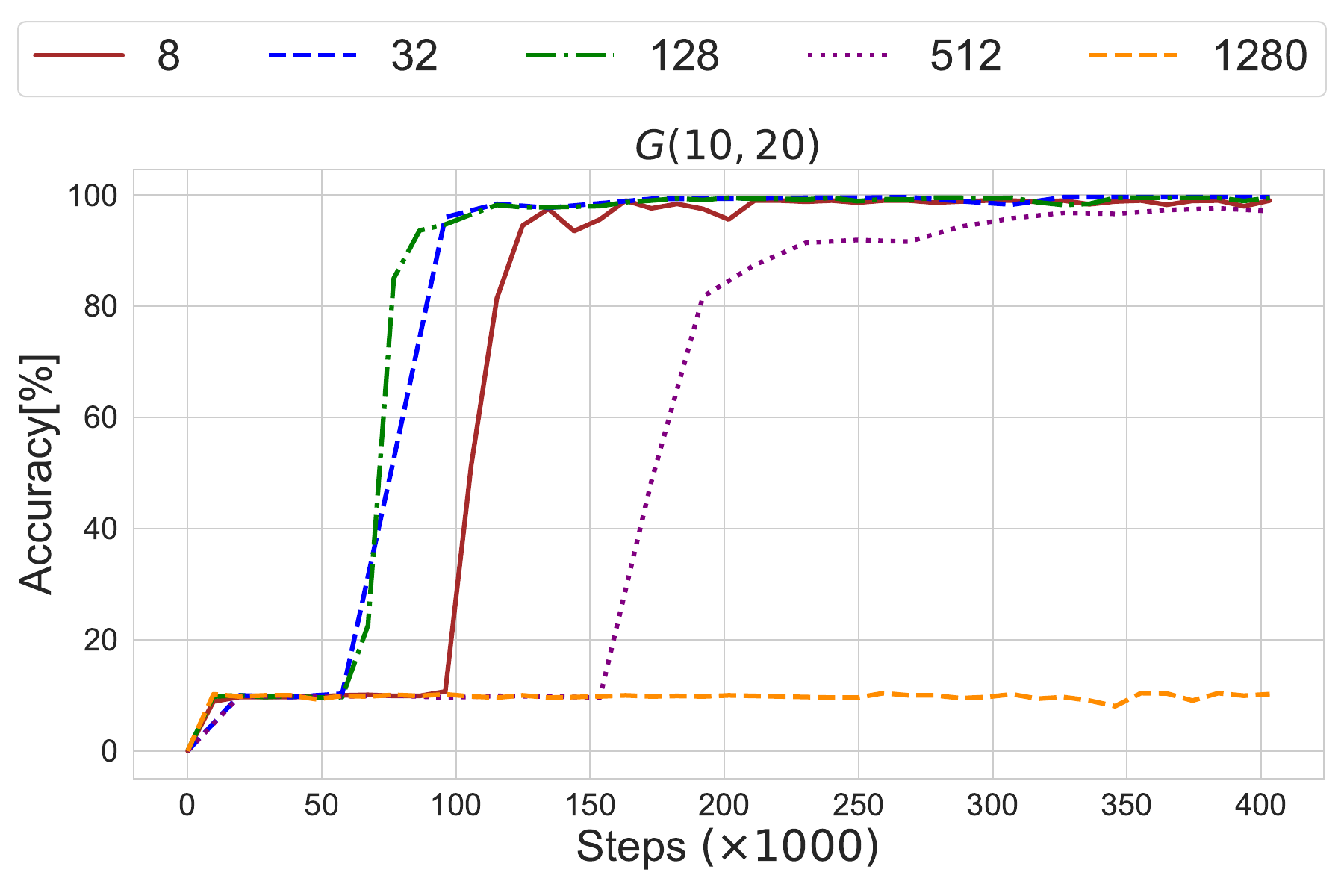}
\caption{
Convergence curves of models with different latent dimensions.
}\label{dim_convergence}
\end{figure}

Figure \ref{dim_convergence} reveals the impact of the latent dimension. 
Dimension reduction helps both the final accuracy and convergence speed, and using relatively lower dimensions such as 32 is more effective than using higher ones. 
Although the model with a latent dimension of 512 successfully performed the task, it requires a significantly longer time to converge. 
When using the same dimension as the model, we remove the linear transformation and this leads to poor performance, indicating the benefits of using compressed representations.

\paragraph{Number of Planning Tokens}

\begin{figure}[t]
\centering
\includegraphics[width=0.9\linewidth]{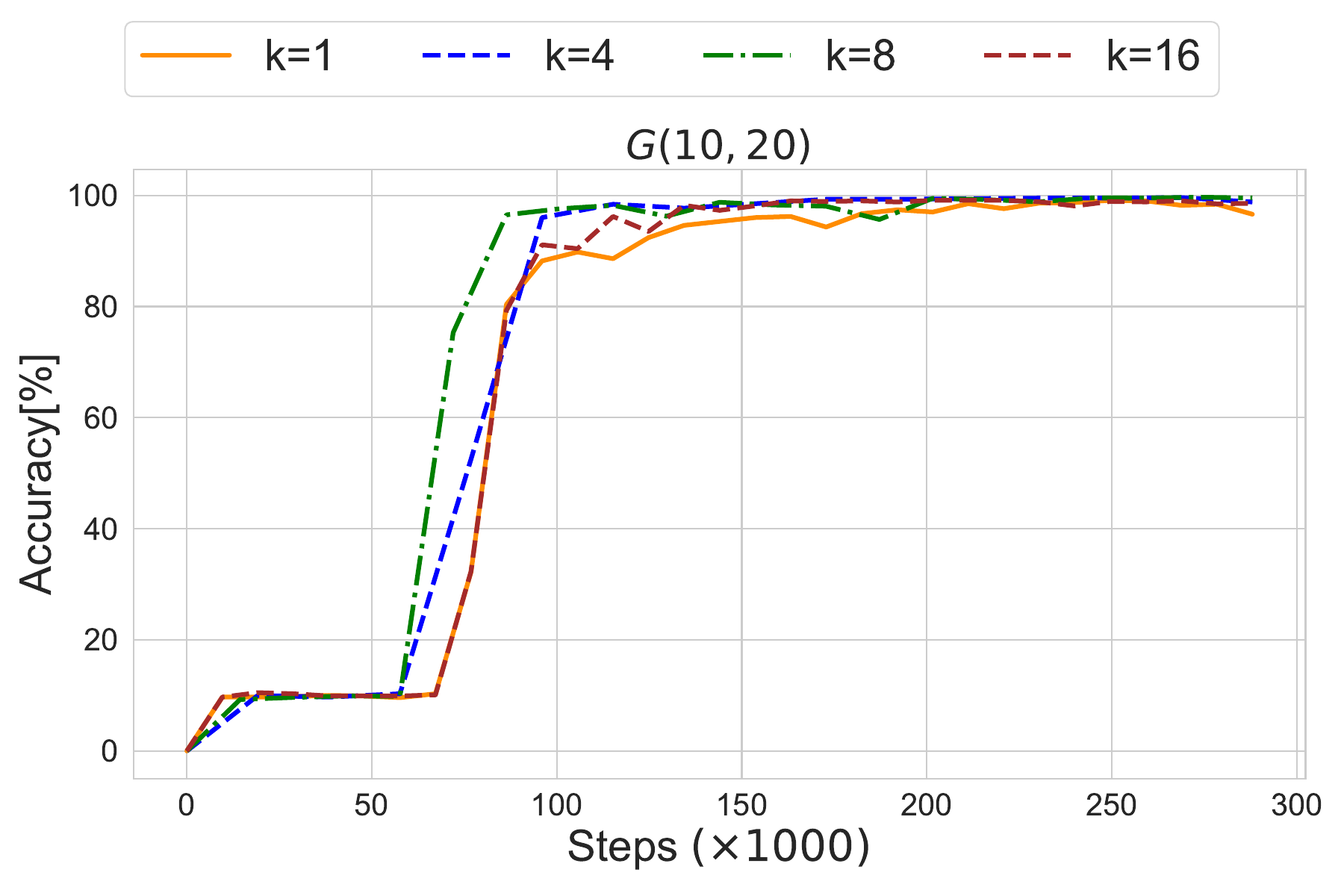}
\caption{
Convergence curves of models with different numbers of planning tokens.
}\label{ztokens_convergence}
\end{figure}

We choose the task setting G(10,20) to examine the effect of the number of planning tokens.
As shown in Figure \ref{ztokens_convergence}, the number of planning tokens does not have a particularly significant impact on the final accuracy. 
This may be because the suffix length is short (<50) and the model capacity is sufficient.
The number of tokens influences the speed of the convergence, and the model converges fastest with $k=8$.

\subsubsection{Attention Visualization}

\begin{figure}[t]
  \centering
  \subfloat[Attention Visualization of Complete Sequence]
  {\includegraphics[width=1.0\linewidth]{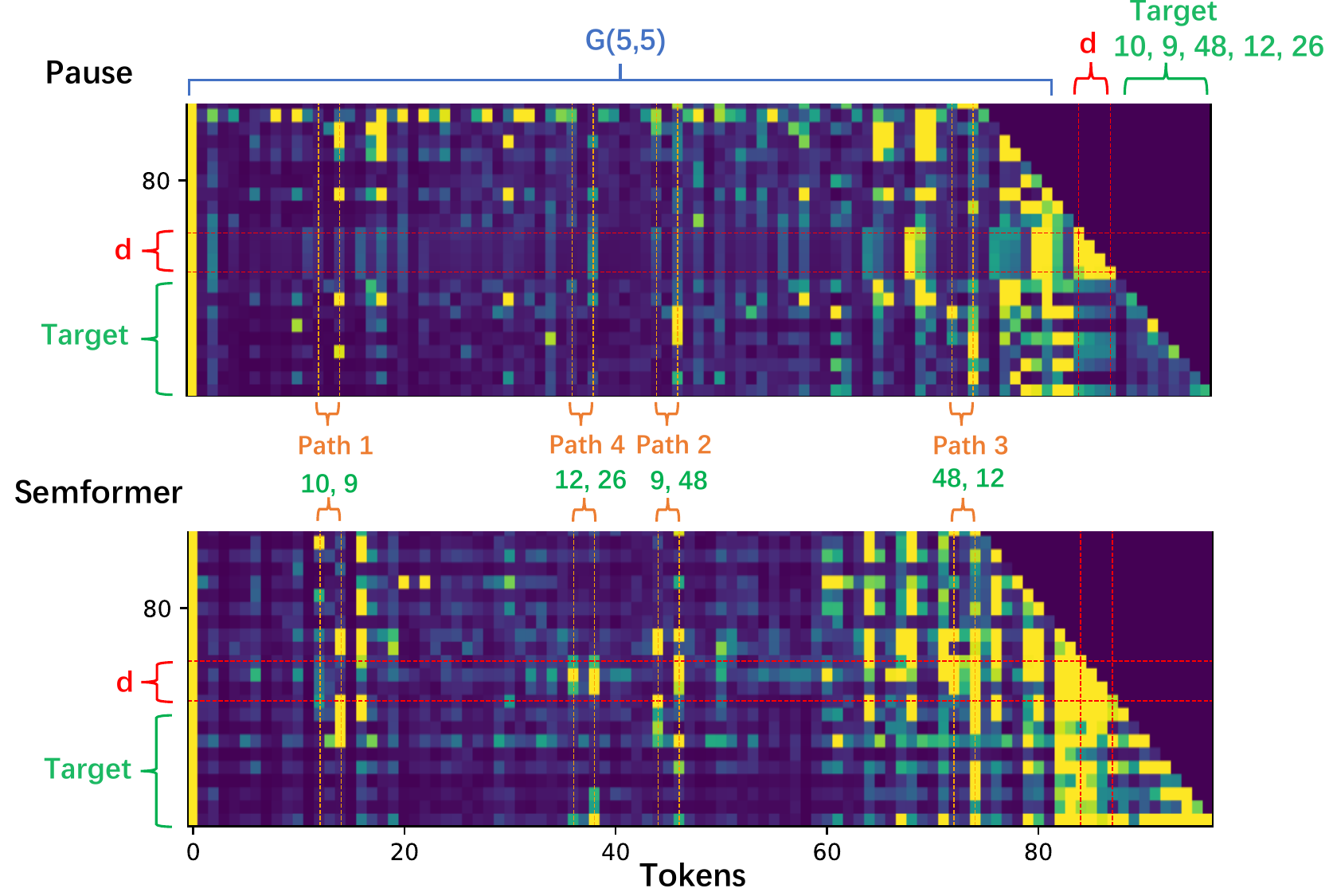}}
  \\
  \subfloat[Attention Visualization of Special Tokens]
  {\includegraphics[width=1.0\linewidth]{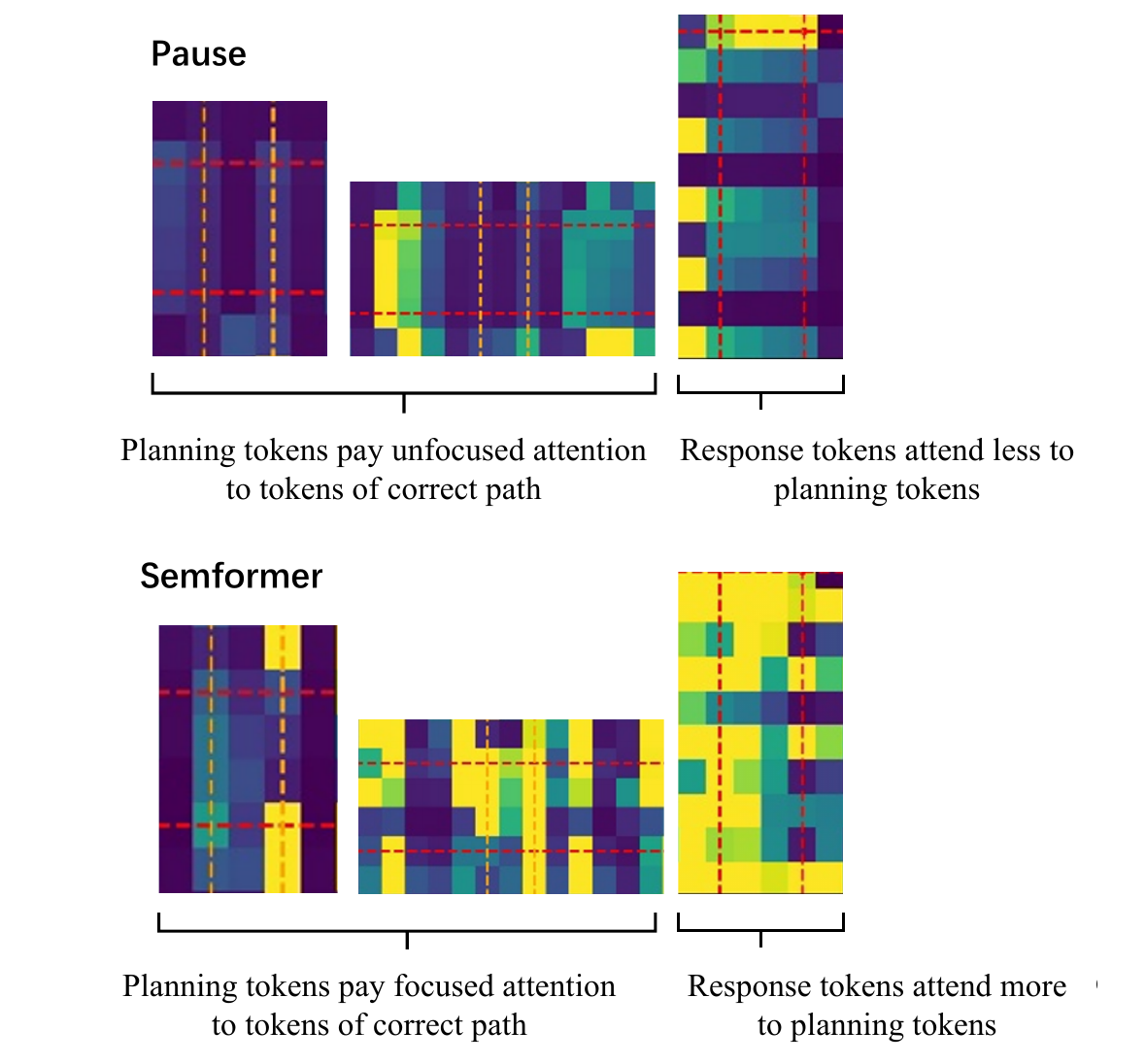}}
\caption{Visualization of Pause and Semformer's attention weights.
}
\label{attenvis}
\end{figure}

We conduct a visualization of the attention weights to see what information is captured by these tokens (Figure \ref{attenvis}).
We select the graph setting G(5,5) and the pretrained model is GPT2-Large.
For each layer, we average the attention weights from all the attention heads, and observe a shift in the attention distribution in the 28th layer of Semformer. 
The planning tokens are successful in capturing the paths leading to the answer. 
Moreover, the answer tokens not only concentrate on their context but also allocate sufficient attention to the planning tokens. 
This contrasts with the Pause model, where the planning tokens fail to capture the correct paths, and the attention from the answer tokens to the planning tokens is insignificant.

\section{Experiments of Pretraining}

In this section, we extend the proposed model to pre-training, and validate its effectiveness in terms of perplexity, in-context learning, and supervised fine-tuning.

\subsection{Setting}
We train a Transformer language model with the same configuration as GPT2, totaling 125M parameters.
The corpus is the public version of OpenWebText. 
We use a sequence length of 1,024, and the batch size is 512.
For each sequence, we randomly split it into a suffix and a prefix, ensuring that the prefix contains at least 128 tokens.
Following \cite{hewitt2023backpack}, we set the gradient steps to 100,000 and it approximately runs 6 epochs. The optimizer is AdamW with a learning rate of 6e-4 and a warmup of 5,000 steps. 
The number of planning tokens and the latent dimension are set to 16 and 64, respectively.
The two numbers are set empirically and we do not tune them.
For the coefficient of the regularization $\alpha$, we select it from $\{0.1,0.5,1.0\}$ according to the perplexity on Wikitext \cite{merity2016pointer}, and find that the model achieves lowest perplexity with $\alpha=0.5$.
In addition to our proposed model, we also train a vanilla Transformer model and a model without latent representation prediction (i.e., Pause) using the identical hyper-parameters.

\subsection{Results}

\begin{table}
\centering
\small
\begin{tabular}{l|c|c}
\toprule
Model & \textbf{Wikitext} & \textbf{LAMBADA} \\
\midrule
TF   &  37.5 & 42.5/32.1  \\
TF-Pause & 35.9 & 43.3/32.7 \\
Semformer & \textbf{35.6} & \textbf{38.8}/\textbf{33.5} \\
\bottomrule
\end{tabular}
\caption{
\label{tab_gpt2_ppl}
Language modeling performance measured by perplexity.
For LAMBADA, we additionally report the accuracy followed by the perplexity score. 
The optimal results are highlighted in bold.
}
\end{table}

\paragraph{Perplexity}
The perplexity scores are shown in Table \ref{tab_gpt2_ppl}.
On the Wikitext test set, we simply insert planning tokens at the middle position of each sequence. 
Different from Wikitext, LAMBADA is dedicated to investigating the long-range dependencies in text \cite{paperno-etal-2016-lambada}, and the perplexity is only calculated on the tokens to predict.
Similarly, Semformer achieves the lowest perplexity due to that the representation prediction encourages the model to predict the whole future semantic representations in advance.
The performance gap between Semformer and TF-Pause has become more significant compared to that on Wikitext.
Moreover, even without the tokens, our model achieves lower perplexity than the other two baselines (35.6 on Wikitext and 39.5 on LAMBADA), indicating that our framework also yields better representation learning.

\paragraph{In-context Learning}

\begin{figure}[htbp]
  \centering
  \subfloat[SST-2]
  {\includegraphics[width=0.4\textwidth]{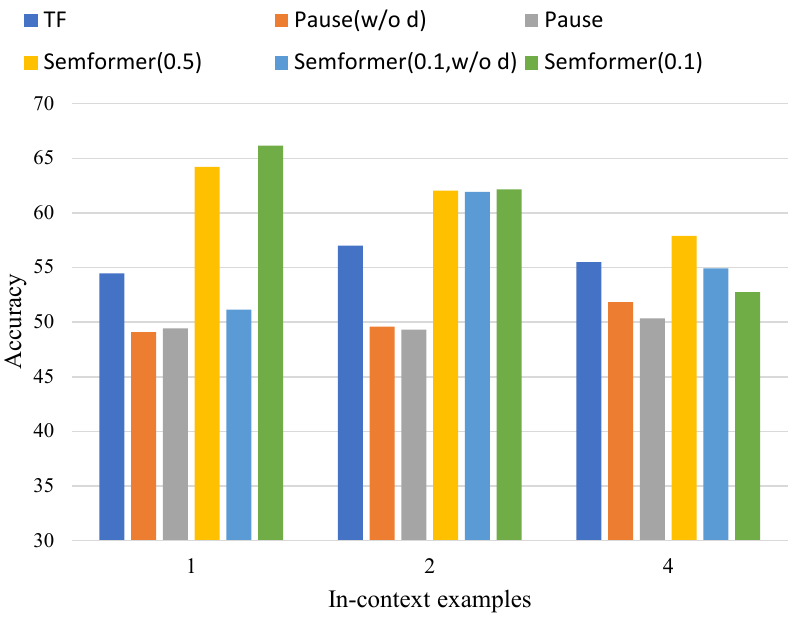}}\\
  \subfloat[MRPC]
  {\includegraphics[width=0.4\textwidth]{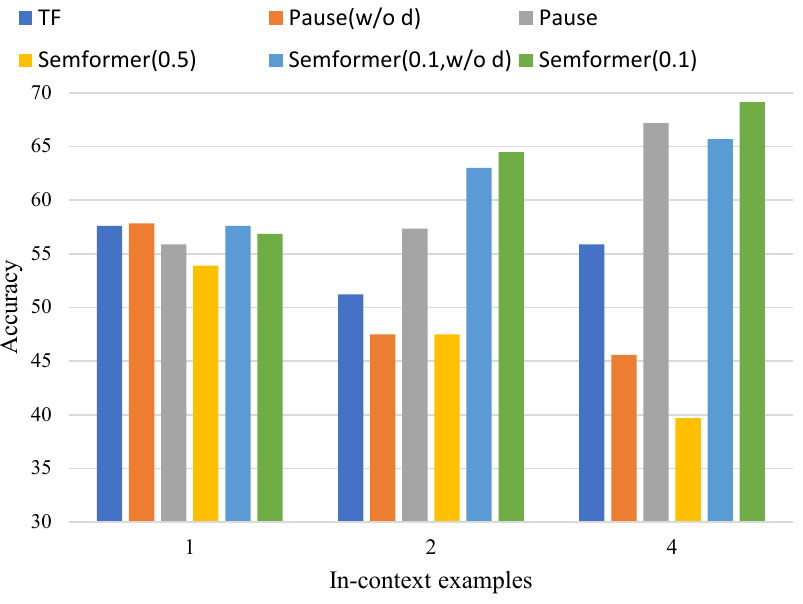}}
  \caption{
In-context learning performance.
}\label{icl_sst}
\end{figure}

We select a single-sentence classification task, Stanford Sentiment Treebank Binary (SST-2) \cite{socher-etal-2013-recursive}, and a paraphrase identification task, Microsoft Research Paraphrase Corpus (MRPC) \cite{dolan-brockett-2005-automatically}, to investigate the performance of in-context learning (ICL)\footnote{https://github.com/EleutherAI/lm-evaluation-harness}. 
The results are presented in Figure \ref{icl_sst}, and we observed the following phenomena.
Semformer performs best on both tasks, achieving an accuracy of 66.1 on SST-2 and 69.1 on MRPC. 
In contrast, the best TF model achieves 57.0 on SST-2 and 57.6 on MRPC.
Specifically, when TF-Pause does not utilize planning tokens during inference, there was a significant decline in performance. 
However, the performance decrease of Semformer is not as pronounced when removing the planning tokens, demonstrating the improvement in representation learning due to the regularization.
Furthermore, a larger coefficient of 0.5 is found inferior to a smaller one, i.e., 0.1. 
This may be because such classification tasks do not heavily rely on lookahead ability, and the model requires a balance between the use of context and the prediction of future information.
Scaling up the model size to increase its capability could potentially mitigate this phenomenon, and we leave this as a future investigation.

\paragraph{Supervised Fine-tuning on Summarization}

\begin{table}[t]
\centering
\small
\begin{tabular}{lccc}
\toprule
Model & R-1 & R-2 & R-L \\
\midrule
\multicolumn{4}{c}{XSum} \\
\midrule
TF   & 35.86 & 13.94 & 28.61 \\
TF-Pause & 35.85 &  13.85 & 28.60 \\
Semformer  & \textbf{36.47} & \textbf{14.37} & \textbf{29.07} \\
\midrule
\multicolumn{4}{c}{SAMSum} \\
\midrule
TF   & 45.60 & 21.09 & 41.62\\
TF-Pause & 46.74 &  21.96 & 42.54 \\
Semformer  & \textbf{46.93} & \textbf{22.29} & \textbf{42.72} \\
\midrule
\multicolumn{4}{c}{DialogSum} \\
\midrule
TF   & 42.65 & 16.54 & 37.50 \\
TF-Pause & 42.17 & 16.47 & 37.09 \\
Semformer  & \textbf{43.18} & \textbf{16.59} & \textbf{38.02} \\
\bottomrule
\end{tabular}
\caption{
\label{tab_summ}
Evaluation on abstractive text summarization.
}
\end{table}

In this section, we investigate the performance of supervised fine-tuning of the whole framework on abstractive summarization.
We use XSum \cite{narayan-etal-2018-dont}, SAMSum \cite{gliwa-etal-2019-samsum}, and DialogSum \cite{DBLP:conf/inlg/ChenLZ21} for evaluation, and report ROUGE-1, ROUGE-2, and ROUGE-L \cite{lin-2004-rouge}. 
We finetune each model on the training data of each task separately and select the checkpoints with highest ROUGE-L score on the individual validation set. 
The batch size is 128 and the learning rate is set to 5e-5.
The value of $\alpha$ is set to 0.5, consistent with its setting during pretraining.
We use beam search with a beam size 2 for all of the models.
The results in Table \ref{tab_summ} show that the Semformer outperforms the standard Transformer and TF-Pause, indicating that the mechanism of semantic planning modeling is beneficial for abstractive summarization.




\section{Conclusion}
In this paper, we presented Semformer which explicitly models semantic planning in addition to next-token prediction.
Semformer introduces a sequence of trainable planning tokens to induce the planning within the internal computation, and the planning tokens in the language model are supervised by predicting the latent planning representations generated by an autoencoder.
The results on the graph path-finding problem show that Semformer can achieve nearly perfect accuracy in such a minimal lookahead task, alleviating the shortcut learning caused by teacher forcing. 
Extending Semformer to a general pertaining on OpenWebtext demonstrates the advantages of the paradigm.

Future research will focus on validating our model with larger sizes and training corpus and exploring its application on reasoning-related tasks such as math and coding. 
Additionally, investigating hierarchical or block-wise prediction of semantic vectors presents a promising avenue for further exploration.

\newpage
\section{Limitations}
Due to limited computation resources, we only pretrain a language model with 125M.
Whether our method can still outperform teacher forcing when combined with larger corpora and when the model size scales up to 1B or even larger needs to be verified in the future. 
In addition, we do not provide theoretical analysis to prove that the method can mitigate the bias in teacher forcing.


\bibliography{anthology,custom}
\bibliographystyle{acl_natbib}

\appendix



\end{CJK}
\end{document}